\documentclass{article}

\usepackage[final]{neurips_2022}

\usepackage[utf8]{inputenc} 
\usepackage[T1]{fontenc}    
\usepackage{hyperref}       
\usepackage{url}            
\usepackage{booktabs}       
\usepackage{amsfonts}       
\usepackage{nicefrac}       
\usepackage{microtype}      
\usepackage{xcolor}         
\usepackage{graphicx}
\usepackage{float} 

\usepackage{amsmath,amsfonts,bm}

\def\eqref#1{equation~\ref{#1}}

\def\1{\bm{1}}

\def\rva{{\mathbf{a}}}

\def\vs{{\bm{s}}}

\def\vx{{\bm{x}}}
\def\vy{{\bm{y}}}

\def\mM{{\bm{M}}}

\DeclareMathAlphabet{\mathsfit}{\encodingdefault}{\sfdefault}{m}{sl}
\SetMathAlphabet{\mathsfit}{bold}{\encodingdefault}{\sfdefault}{bx}{n}

\newcommand{\E}{\mathbb{E}}

\newcommand{\R}{\mathbb{R}}

\title{Differentiable Neural Computers \\with Memory Demon}

\author{%
  Ari Azarafrooz\\
  \texttt{ari.azarafrooz@gmail.com} \\
}

\begin{document}

\maketitle

\begin{abstract}
 A Differentiable Neural Computer (DNC) \cite{DNC} is a neural network with an external memory which allows for iterative content modification via read, write and delete operations.
 We show that information theoretic properties of the memory contents play an important role in the performance of such architectures. We introduce a novel concept of \textit{memory demon} \footnote {The name is inspired by the concept of Maxwell's Demon who decreases the \textit{entropy} of gas in a box by letting all the high-velocity molecules accumulate on one side and all the low-velocity ones on the other.} to DNC architectures which modifies the memory contents implicitly via additive input encoding. The goal of the memory demon is to maximize the expected sum of mutual information of the consecutive external memory contents. 
 
Github codes \href{here}{https://github.com/azarafrooz/dnc-with-demon}
 
 
  \end{abstract}

\section{Introduction}
A Differentiable Neural Computer (DNC) \cite{DNC} is a neural network coupled to an external memory matrix $\mM \in \R^{N \times W}$ where $W$ is the word/cell length and $N$ is the number of cells and independent of the number of training parameters.  Previous researches have shown external memory matrix provide proper architectural bias for solving algorithmic and structured tasks.

The neural network in DNC is referred to as the `controller'. It interacts with the external world via input $\vx_t \in \R$ and outputs $\vy$.  But it also interacts with the memory matrix at each time step $\mM_t$ using read and write heads to read from $\mM_t$ and to arrive at the next memory matrix $\mM_{t+1}$.  

One main mechanism to allow for the interaction between heads and the memory is content-based addressing. In this mechanism, a key vector is emitted by the controller as an approximation to a part of the stored data which is then compared to memory $\mM$ to yield the exact value. The comparison metric is a cosine similarity score that determines weightings that should be used by the read headers or by the write header to modify the memory content.  


\textbf{Keep memories interesting}
 A technique common between Memory Champions is to assign interesting imageries to the subjects of memorization, also known as the memory palace \cite{Moonwalk}. For example, a certain number is an interesting event or picture. Interesting imageries seem to make the memorization/recall easier. Keeping memory associations interesting, seems to be an efficient strategy to reduce the interference of current and past information and avoids the memory contents to be overwritten or misinterpreted. Inspired by such a technique and in the context of DNC, we setup a Reinforcement Learning (RL) framework where a RL agent is equipped with a mutual information-based reward and an encoding action. Mutual information-based reward $I(\mM_t; \mM_{t+1})=H(\mM_t)+H(\mM_{t+1})-H(\mM_t,\mM_{t+1})$ serves as a proxy to measure ``interestingness''. The intuition is that, if the DNC dynamics are ``too simple'', then $I$ will be small because both entropy terms $H(\mM_t)$ and $H(\mM_{t+1})$ are small. On the other hand, if the DNC dynamics are ``too random'', then $I$  will be small because $H(\mM_t,\mM_{t+1}) \approx H(\mM_t)+H(\mM_{t+1})$. As a result, ``interesting'' non-randomness will exist only in the intermediate regime with high value of $I(\mM_t; \mM_{t+1})$. 
The RL agent action is to assign to each input embedding a proper encoding, akin to assigning a certain imagery for each subject of memorization.

\section{DNC with Memory demon}
At each time step, demon modifies samples via continuous embedding $\rva_t \sim \pi(\vs_t)$ which gets added to the input data $\vx_t$ where $\vs_t = (\vx_t, \mM_t) $.  As a result, the input of the controller is $\vx_t+\rva_t$ which leads to the writing the next memory content $\mM_{t+1}$.
One can view $\rva_t$ as interestingness encoding (rather than a positional encoding.) This iterative process is visualized in figure 1.

Demon then computes the mutual information of the consecutive memory contents $I(\mM_t; \mM_{t+1})$ as its reward and updates its policy $\pi$ in order maximize the expected sum of rewards  $\sum_t \E_{(\vs_t, \rva_t) \sim \rho_{\pi}} I(\mM_t;\mM_{t+1})$.

In order to estimate mutual information of consecutive memory contents, we utilize \cite{Belghazi2018MINEMI}. RL agent then uses this measure to guide its policy encoding actions using PPO \cite{Schulman2017ProximalPO}. 

\begin{figure}[H]
\begin{center}
\includegraphics[width=0.65\textwidth]{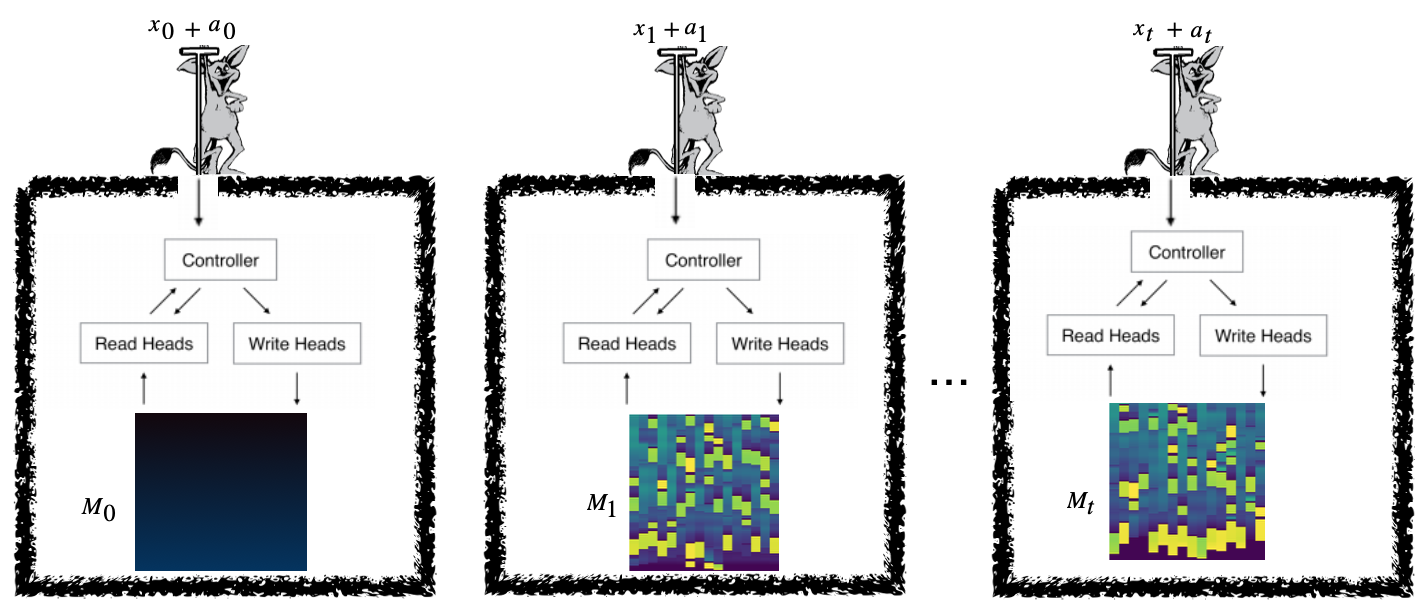}
\end{center}
\caption{Memory Demon encodes the observations with additive encodings $\rva_t$ and receives the mutual information of the consecutive memory contents $I(\mM_t; \mM_{t+1})$ as reward.} 
\end{figure}

\section{Results}

We refer to our architecture as Demon-DNC. 
The most important observation was that we found out that Demon-DNC works best when combined with \textit{(key) masking techniques} of \cite{Csords2019ImprovingDN}. One known issue with content-based memory addressing is that the entire key and cell values are being used to produce the similarity score. This can flatten the memory address distribution when the unknown part of the cell values are more significant. The dynamic and high variance nature of RL agent policies seem to exacerbate this situation which in turn makes the role of masking more important. 

To measure the efficacy of the Demon-DNC, we utilize the 3 following common tasks in the \cite{Csords2019ImprovingDN}. Please refer to the description of such tasks to \cite{Csords2019ImprovingDN} for a more comprehensive description.

\subsection{Associated-recall and Copy task}
\begin{figure}[h] 
	\begin{minipage}[t]{4cm} 
		\centering 
		\includegraphics[scale=0.4]{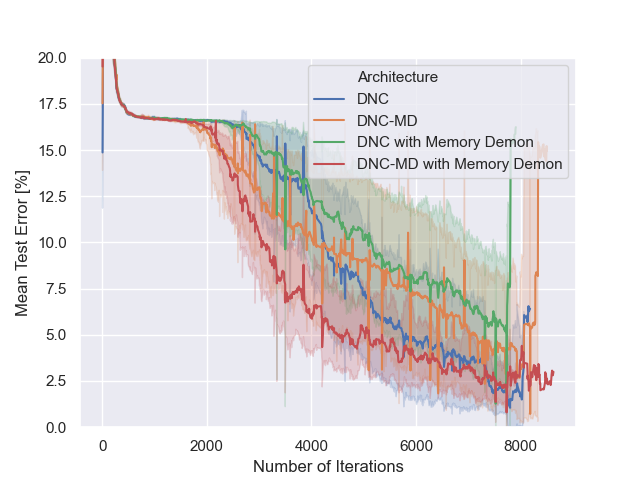} 
		\caption{\tiny Effect of memory demon on the convergence: Associative Recall  } 
	\end{minipage} 
	\hspace{3cm} 
	\begin{minipage}[t]{4cm} 
		\centering 
		\includegraphics[scale=0.4]{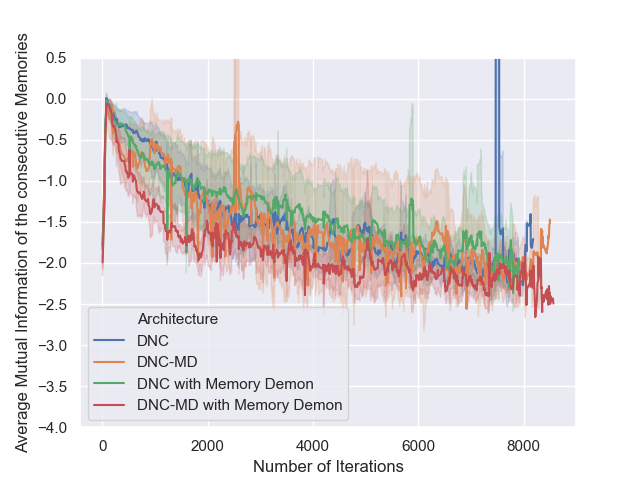} 
		\caption{\tiny Effect of memory demon on the mutual information of consecutive memories: Associative Recall} 
	\end{minipage} 

	\begin{minipage}[t]{4cm} 
		\centering 
		\includegraphics[scale=0.4]{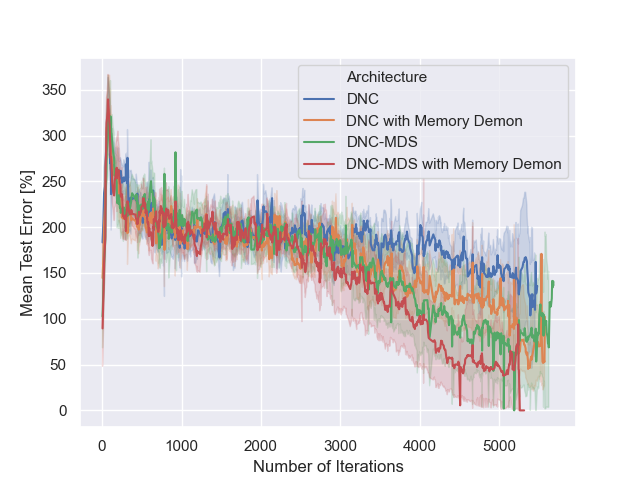} 
		\caption{\tiny Effect of memory demon on the convergence: Repeat-Copy } 
	\end{minipage} 
	\hspace{3cm} 
	\begin{minipage}[t]{4cm} 
		\centering 
		\includegraphics[scale=0.4]{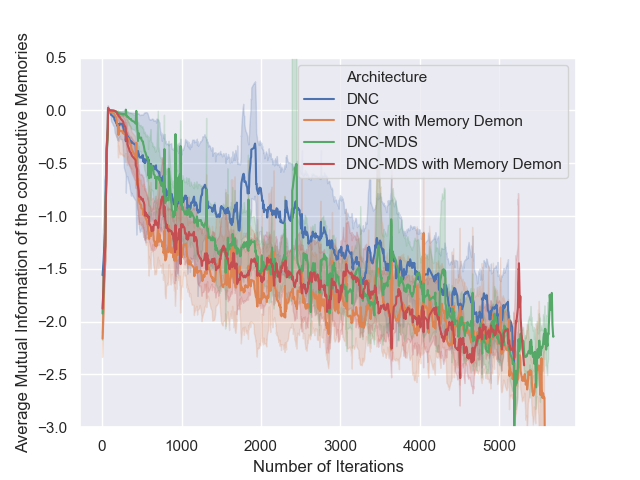} 
		\caption{\tiny Effect of memory demon on the mutual information of consecutive memories: Repeat-Copy } 
	\end{minipage} 
 
\caption{Associative Recall and Repeat-Copy experiment over 10 seeds.}
\end{figure} 
Fig. 6 shows that Demo-DNC with masking is more efficient than the rest in both tasks. Although the improvement (in terms of the convergence) seems to be more significant for the Associated-recall task than for the copy task. Moreover, it shows the drops in the mutual information loss measures are consistent with drop in the mean test errors. Networks that have lower mean test errors have also lower mutual information loss, confirming our main assertion on the role of information theoretic properties of the memory contents.  

\subsection{bAbi-10 task}

We also conducted the bAbI experiments and the results are reported in Table \ref{table:babi-10k}. As it can be seen, given the same number of training steps, not only Demon-DNC outperforms the best DNC network in mean, it also outperform it in 9 tasks.

\begin{table}
  \caption{10K-bAbI error rates of different models after 0.5M iterations of training [\%]}
  \label{table:babi-10k}
  \centering
  \begin{tabular}{lllllll}
    \toprule
    Task  & DNC  & DNC-MDS & DNC-DS & DNC-MS & DNC-MD & DNC-MD-Demon\\
    \midrule
    1 & $2.5\pm4.4$ &  $0.4\pm1.2$  & $0.7\pm1.6$  & $0.0\pm0.1$ & $\bm{0.0\pm0.0}$ & $\bm{0.0\pm 0.0}$ \\
    2 & $29.0\pm19.4$ & $8.6\pm10.1$ & $18.6\pm15.1$ & $7.8\pm5.9$ & $6.9\pm4.7$ & $\bm{1.54\pm 0.57}$  \\
    3 & $32.3\pm14.7$ & $10.8\pm9.5$ & $16.9\pm13.0$ & $\bm{7.9\pm7.8}$ & $12.4\pm5.1$ & $8.34 \pm 0.81$ \\
    4 & $0.8\pm1.5$ & $0.8\pm1.5$ & $6.4\pm10.0$ & $0.8\pm1.0$ & \bm{$0.1\pm0.2$} & $0.2\pm0.3$ \\
    5 & $1.5\pm0.6$ & $1.6\pm1.0$ & $1.3\pm0.5$ & $1.7\pm1.1$ & $1.3\pm0.7$ & $\bm{0.97\pm0.17}$\\
    6 & $5.2\pm6.8$ & $1.1\pm2.1$ & $2.4\pm3.8$ & $\bm{0.0\pm0.1}$ & $0.1\pm0.1$ & $0.1\pm0.1$\\
    7 & $8.8\pm5.8$ & $3.4\pm2.3$ & $7.6\pm5.1$ & $2.5\pm2.0$ & $3.0\pm5.0$ & $\bm{1.24\pm1.04}$ \\
    8 & $11.6\pm9.4$ & $4.6\pm4.5$ & $10.9\pm7.9$ & $\bm{1.8\pm1.6}$ & $2.5\pm2.1$ & $1.91\pm 1.19$ \\
    9 & $4.5\pm5.8$ & $0.8\pm1.9$ & $2.0\pm3.3$ & $\bm{0.1\pm0.2}$ & $0.1\pm0.2$ & $ \bm{0.06 \pm 0.06}$ \\
    10 & $9.1\pm11.5$ & $2.6\pm3.9$ & $4.1\pm5.9$ & $0.6\pm0.6$ & $0.5\pm0.5$ & $ \bm{0.33  \pm 0.27}$ \\
    11 & $11.6\pm9.4$ & $0.1\pm0.1$ & $0.1\pm0.2$ & $\bm{0.0\pm0.0}$ & $\bm{0.0\pm0.0}$ & $\bm{0.0 \pm 0.0}$\\
    12 & $1.1\pm0.8$ & $\bm{0.2\pm0.2}$ & $0.5\pm0.4$ & $0.3\pm0.4$ &  $0.2\pm0.2$ & $\bm{0.03 \pm 0.07}$ \\
    13 & $1.1\pm0.8$ & $\bm{0.1\pm0.1}$ & $0.2\pm0.2$ & $0.2\pm0.2$ & $\bm{0.1\pm0.1}$ & $ 0.16 \pm 0.34$\\
    14 & $24.8\pm22.5$ & $8.0\pm13.1$ & $20.0\pm19.4$ & $1.8\pm0.9$ & $ 2.0\pm1.6 $ & $ \bm{1.97 \pm 0.73}$ \\
    15 & $40.8\pm1.4$ & $26.3\pm20.7$ & $42.1\pm6.3$ & $33.0\pm15.1$ & $23.6\pm18.6$ & $ \bm{6.46 \pm 12.32}$ \\
    16 & $\bm{53.1\pm1.2}$ & $54.5\pm1.8$ & $53.5\pm1.4$ & $53.2\pm2.3$ & $53.9\pm1.2$ & $53.28 \pm 2.18$ \\
    17 & $\bm{37.8\pm2.5}$ & $39.9\pm3.2$ & $40.1\pm2.0$ & $41.2\pm3.0$ & $39.8\pm1.2$ & $ 39.14 \pm 1.89$ \\
    18 & $7.0\pm3.0$ & $6.3\pm4.1$ & $9.4\pm0.9$ & $3.3\pm2.2$ & $2.0\pm2.6$ & $\bm{1.67 \pm 1.64}$ \\
    19 & $67.6\pm8.6$ & $48.6\pm32.8$ & $67.6\pm7.9$ & $48.1\pm26.7$ & $40.7\pm34.9$ & $ \bm{14.08 \pm 2.87}$\\
    20 &  $\bm{0.0\pm0.0}$ & $0.9\pm0.9$ & $1.5\pm1.0$ & $5.3\pm12.5$ & $0.1\pm0.1$ & $ 0.06 \pm  0.06$\\
    mean & $16.9\pm5.2$ & $11.0\pm3.8$ & $15.3\pm3.5$ & $10.5\pm1.9$ &$9.5\pm1.6$ & $ \bm{6.58 \pm 0.73}$ \\
    \bottomrule
  \end{tabular}
\end{table}

\begin{table}
  \caption{1K-bAbI error rates of different models}
  \label{table:babi-1k}
  \centering
  \begin{tabular}{lll}
    \toprule
    Model  & Error\\
    \midrule
     DNC-MD-Demon & $9.98$  $\bm{(7/20})$ \\
    Universal Transformer (UT) &  $\bm{8.50}$ $(8/20)$ \\
    \bottomrule
  \end{tabular}
\end{table}

\begin{table}
  \caption{1K-bAbI error rates of different models across tasks}
  \label{table:babi-1k-h}
  \centering
  \begin{tabular}{lll}
    \toprule
    Task  & UT with dynamic halting & DNC-MD-Demon \\
    \midrule
    1 & $\bm{0.0}$ &  $\bm{0.0$} \\
    2 & $\bm{0.5}$ & $27.34$ \\
    3 & $\bm{5.4}$&  $24.12$  \\
    4 &  $\bm{0.0}$ & $2.2$  \\
    5 &  $\bm{0.5}$ & $0.8$  \\
    6 &  $\bm{0.5}$ &  $0.7$  \\
    7 &  $\bm{3.2}$ &  $7.54$  \\
    8 & $\bm{1.6}$ &  $2.16$ \\
    9 &  $\bm{0.2}$ &  $0.3$  \\
    10 & $\bm{0.4}$ & $1.21$ \\
    11 & $\bm{0.1}$ & $\bm{0.1}$ \\
    12 & $\bm{0.0}$ & $\bm{0.0}$ \\
    13 & $0.6$ & $\bm{0.2}$ \\
    14 & $3.8$ & $\bm{0.6}$ \\
    15 & $5.9$ & $\bm{0.0}$ \\
    16 & $\bm{15.4}$ & $53.75$ \\
    17 & $42.9$ & $\bm{28.63}$ \\
    18 & $\bm{4.1}$ & $5.62$ \\
    19 & $68.2$ & $\bm{44.24}$ \\
    20 & $2.4$ & $\bm{0.0}$ \\
  \bottomrule
  \end{tabular}
\end{table}

\section{Shortcomming and future work}

\subsection{Comparing with some of the state of art}
We also experimented on bAbi-1K.  Since we found no available numbers on previous DNC-networks, we only compare it with UT \cite{Dehghani2019UniversalT} in Table \ref{table:babi-1k} and Table \ref{table:babi-1k-h}. It can be seen that after 0.5M iterations, UT outperforms Demon-DNC. However, the full potential of external memories might have not been explored fully. The idea of optimizing the information theoretic properties of memory contents (with or without RL agents) for example might inspire other fresh ideas for networks with external memory.

\subsection{Scalability}  
One attractive properties of DNC is that number of \textit{parameters} is independent of the number of memory cells $N$. However, this is not true for the information estimator and RL networks. We are exploring the feasibility of integrating our framework with Sparse DNC \cite{Rae2016ScalingMN} to address this issue.



\bibliographystyle{plainnat}
\bibliography{neurips_2022}

\end{document}